# Zero Memory Overhead Approach for Protecting Vision Transformer Parameters against Bit-flip Faults

Fereshteh Baradaran
School of Electrical and Computer
Engineering
Shiraz University
Shiraz, Iran
f.baradaran@hafez.shirazu.ac.ir

Mohsen Raji
School of Electrical and Computer
Engineering
Shiraz University
Shiraz, Iran
mraji@shirazu.ac.ir

Azadeh Baradaran
School of Electrical and Computer
Engineering
Shiraz University
Shiraz, Iran
azd.baradaran@gmail.com

Arezoo Baradaran
School of Electrical and Computer
Engineering
Shiraz University
Shiraz, Iran
arzo.baradaran@gmail.com

Reihaneh Akbarifard
School of Electrical and Computer
Engineering
Shiraz University
Shiraz, Iran
Reihanehakbarifard@gmail.com

*Abstract*—**Vision Transformers (ViTs) have demonstrated superior performance over Convolutional Neural Networks (CNNs) in various vision-related tasks such as classification, object detection, and segmentation due to their use of self-attention mechanisms. As ViTs become more popular in safety-critical applications like autonomous driving, ensuring their correct functionality becomes essential, especially in the presence of bit-flip faults in their parameters stored in memory. In this paper, a fault tolerance technique is introduced to protect ViT parameters against bit-flip faults with zero memory overhead. Since the least significant bits of parameters are not critical for model accuracy, replacing the LSB with a parity bit provides an error detection mechanism without imposing any overhead on the model. When faults are detected, affected parameters are masked by zeroing out, as most parameters in ViT models are near zero, effectively preventing accuracy degradation. This approach enhances reliability across ViT models, improving the robustness of parameters to bit-flips by up to three orders of magnitude, making it an effective zero-overhead solution for fault tolerance in critical applications.**

## I. INTRODUCTION

Transformers were initially introduced for natural language processing (NLP) tasks, but their architecture has since been adapted for computer vision, leading to the development of Vision Transformers (ViTs) [1]. ViTs have been highly successful in tasks such as image classification [1], object detection [2], and segmentation [3], significantly outperforming Convolutional Neural Networks (CNNs) [4]. This superior performance has drawn considerable attention from researchers and industry, enabling their application across various fields, including safety-critical domains like autonomous driving [5]. In these environments, hardware faults can impact system performance, making the need for efficient fault-tolerant techniques increasingly important.

Many studies have improved fault tolerance in CNNs, utilizing techniques like range-based bounds checking [6], redundant layer computation [7], and algorithm-based fault tolerance (ABFT) [8]. While these methods enhance resilience in CNNs, they are either insufficient or pose significant overhead for ViTs. In [9], Approximate ABFT is proposed to provide fault-tolerance in ViTs, focusing on non-linear functions in ViTs. A vulnerability-guided suppression is introduced in [10], which zeros out errors in fully connected layers. In [11], ALBERTA is developed, a checksum-based approach for GEMM layers. Existing solutions, however, typically introduce additional memory overhead, which limits their suitability for memory-constrained applications. This gap highlights the need for a lightweight, memory-efficient approach.

This paper presents a fault tolerance technique for protecting parameters of ViT against bit-flip faults with no additional memory overhead. The vulnerability of individual bits in the model's parameters is evaluated, revealing that the least significant bits (LSBs) have minimal impact on model accuracy. Leveraging this insight, the LSB is substituted with a parity bit, offering a lightweight error detection mechanism. After the fault is detected by finding parity mismatches, the affected parameter is masked by replacing it with zero, effectively preventing the error from being propagated in the model. This method maintains model reliability without adding any memory overhead, making it an efficient solution for use in safety-critical applications. The proposed method is evaluated across three ViT models (ViT, DeiT, and Swin) in three versions each (tiny, small, and base). Experimental results demonstrate that the proposed approach enhances the robustness of parameters to bit-flips by up to three orders of magnitude, effectively maintaining reliability across all ViT models. These findings underscore its potential as a zero-overhead solution for fault tolerance in applications requiring dependable performance.

The rest of this paper is organized as follows: Section II provides background on ViTs and the fault model. Section III presents the proposed method for enhancing fault tolerance in ViTs. Section IV discusses experimental results that demonstrate the effectiveness of the proposed approach, and Section V concludes the paper.

## II. BACKGROUND

This section provides an overview of ViTs, including key models like ViT, Swin, and DeiT, and their core components.

It also introduces the fault model, focusing on bit flips in model parameters that impact reliability.

### A. Vision Transformers

Transformers were initially designed for NLP tasks, where they excelled at capturing relationships between words in sequences. This success prompted researchers to adapt transformer architectures for computer vision applications, including image classification, object detection, and segmentation [1]. The ViT architecture consists of multiple transformer encoders, which include essential components like multi-head attention modules (MHA), feed-forward networks (FF), residual connections, and layer normalization. MHA can be formalized as (1) where $d_k$ stands for the dimension of query and key vectors. Unlike CNNs, which process images as whole units, ViTs first divide input images into non-overlapping patches, which are then linearly projected into a sequence for analysis.

$$Attention(Q, K, V) = Softmax(\frac{QK^T}{\sqrt{d_k}})V. \qquad (1)$$

Recently, different models of ViTs have been introduced. The original ViT model showed that pure transformer architectures can outperform leading CNNs on common vision tasks when trained with large datasets [1]. The Swin Transformer features a hierarchical design that uses shifted windows to enable flexible modeling at different scales while keeping computational complexity manageable for image sizes [2]. Additionally, the Data-efficient Image Transformer (DeiT) explored various training strategies to enhance learning performance on smaller datasets [12].

### B. Fault Model

In this study, the focus is on bit flips that occur in the parameters of models. These bit-flips can be caused by various factors, such as radiation-induced soft errors or timing and retention errors related to the memory used for parameter storage [13].

## III. PROPOSED METHOD

This section outlines the proposed method to enhance the reliability of ViTs. It begins with the motivation behind the approach, followed by an introduction to the fault detection mechanism for identifying parameter errors. Lastly, the fault mitigation mechanism is discussed, detailing how these faults are addressed.

### A. Motivation

In order to evaluate the criticality of different bit positions with respect to their impact on the model accuracy, a study is conducted to compute the Bit Error Rate with Zero Accuracy Drop(BERZAD) for different bit positions. The BERZAD metric, introduced by Sabbagh [14], represents the highest bit-error rate (the ratio of erroneous parameter bits to the total number of parameter bits) that maintains zero loss of accuracy, determined by a 95% confidence interval. A higher BERZAD value signifies greater robustness of the network.

Fig. 1 shows the BERZAD for different bit positions for different ViT models. As the figure shows, the LSBs in ViTs have a minimal effect on model accuracy; i.e. flipping a small number of these bits does not significantly impact performance. Leveraging this insight, the concept from [15] is extended, applying a parity-based error detection technique at the parameter level of ViT models.

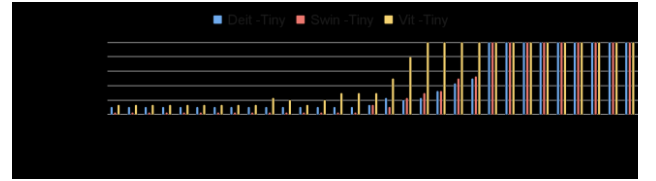

Fig. 1. BERZAD Across Bit Positions for DeiT, Swin, and ViT Models.

### B. Fault Detection

The parity-based error detection technique, originally developed for CNNs called Opportunistic Parity (OP) [15], is adapted to ViT architectures. By applying parity at the parameter level, an odd number of errors can be detected with minimal memory and computational overhead.

To implement this method, even parity is enforced across all parameters. If a parameter has odd parity (i.e., an odd number of 1's in its binary representation), the LSB is flipped to ensure even parity, as shown in Fig. 2. This modification enables the detection of parity mismatches, signaling a potential odd number of faults.

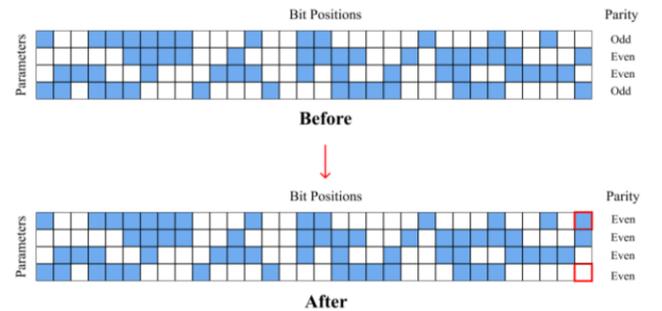

Fig. 2. Fault Detection: Ensuring even parity across all parameters by flipping the LSB.

Since no additional bit is considered for parameter representation and changing the LSB has minimal impact on model accuracy, as illustrated in Fig. 1, this approach provides a zero overhead solution for fault detection with minimal impact on the model performance.

### A. Fault Mitigation

While parity effectively detects an odd number of errors, its primary function is to mitigate the effects of critical errors. Upon detecting a parity error, the affected parameter is masked by setting its value to zero, as illustrated in Fig. 3. This fault-masking approach reduces the impact of flipped Most Significant Bits (MSBs), which can significantly increase the absolute values of the parameters. The distribution of parameters across ViT models, as illustrated by the example of ViT Tiny in Fig. 4, shows that most parameters are near zero. Consequently, any increase in these values due to faults in the MSBs can severely affect model accuracy. By zeroing out erroneous parameters, we can mitigate this adverse effect and prevent a significant drop in accuracy.

| Network | Num. Parameters | Top-1 Accuracy |
|---------|-----------------|----------------|
| Swin-Base | 88 M | 96.17% |

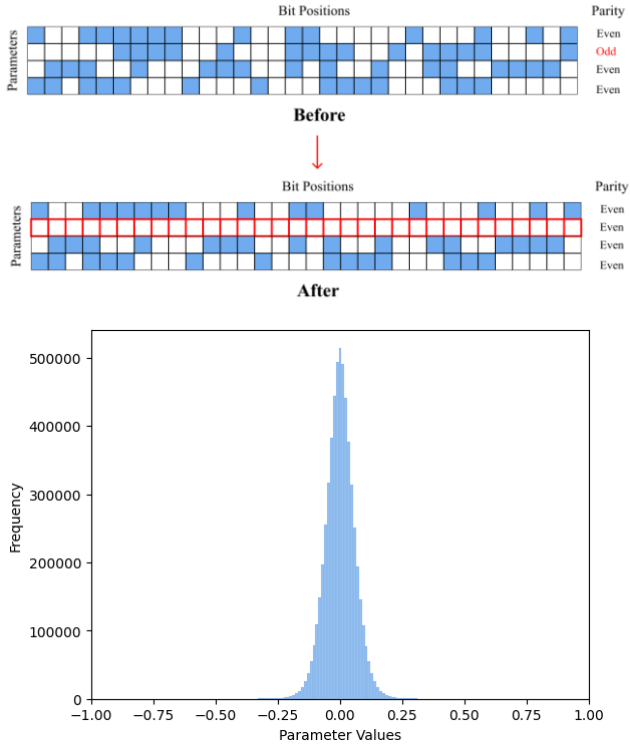

Fig. 4. Distribution of Parameters in ViT-Tiny.

## IV. EXPERIMENTAL RESULTS

In order to show the efficacy of the proposed fault tolerance technique, a set of extensive experiments are conducted on different ViT models. In the following, the experimental setup is explained and then, an effectiveness evaluation of the proposed technique is investigated across various ViT models.

### A. Experimental Setup

To evaluate the robustness of the proposed fault tolerance approach, experiments are conducted on multiple ViT models: ViT, DEiT, and Swin. Each model is tested in three versions—tiny, small, and base. Table 1 summarizes the model architectures, including the number of parameters and their top-1 accuracy on the CIFAR-10 dataset.

TABLE I. CHARACTERISTICS AND ACCURACY OF SELECTED NETWORKS FOR CIFAR10 DATASET

| Network | Num. Parameters | Top-1 Accuracy |
|---------|-----------------|----------------|
| Deit-Tiny | 5.7 M | 93.74% |
| Deit-Small | 22.1 M | 95.55% |
| Deit-Base | 86 M | 95.81% |
| Vit-Tiny | 5.7 M | 95.31% |
| Vit-Small | 22.1 M | 96.33% |
| Vit-Base | 86 M | 93.36% |
| Swin-Tiny | 28 M | 96.43% |
| Swin-Small | 50 M | 96.66% |

The models are evaluated under different Bit Error Rates (BERs), with values ranging from 1e-9 to 1e-1, to simulate varying fault conditions. At each BER level, accuracy is measured both with and without the application of OP, allowing for comparing the effectiveness of the proposed protection method in mitigating errors.

### B. Fault Injection Procedure

To assess the effectiveness of the proposed fault detection and mitigation strategy, a set of fault injection experiments is conducted on various ViT models, as described in Algorithm 1. Random bit flips are introduced into model parameters, both with and without OP protection, by randomly selecting parameters and flipping a single bit in their 32-bit floating-point representation based on a predefined BER.

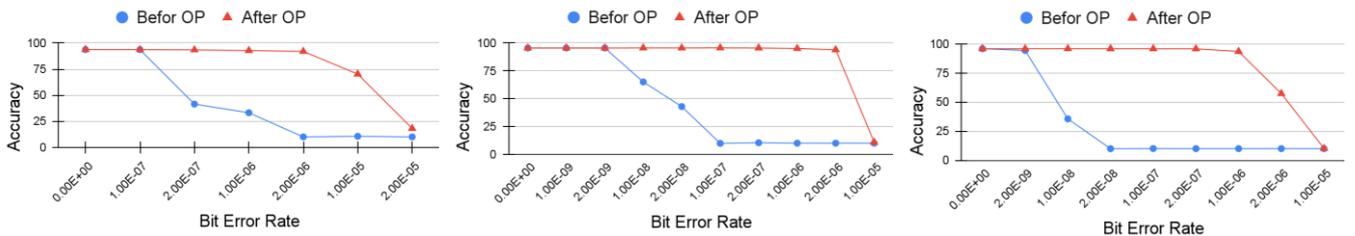

To avoid persistent NaN or infinity values, which are common in fault injection campaigns [14], bit flips are restricted from affecting the MSB of the exponent (bit 30). For statistical accuracy, the fault injection experiments are repeated multiple times, with sample sizes dynamically adjusted to ensure a 95% confidence level in the final accuracy measurements.

### C. Overall Evaluation

Fig. 5 illustrates the accuracy of ViT models as a function of the BER. The blue curves represent the behavior of the models in the absence of any protection, showing a sharp drop in accuracy. This abrupt decline is primarily attributed to the modification of one or a few exponent bits, which drastically impacts the model's performance. In contrast, the red curves representing OP illustrate a much more gradual decline in model accuracy, effectively mitigating extreme cases of accuracy loss and ensuring reliability up to a BER of approximately 1e-5 across all model variants. Interestingly, the Swin model demonstrates higher vulnerability to bit flip faults, with a more abrupt drop in accuracy compared to ViT and DeiT. This increased susceptibility may be attributed to Swin's unique architectural features.

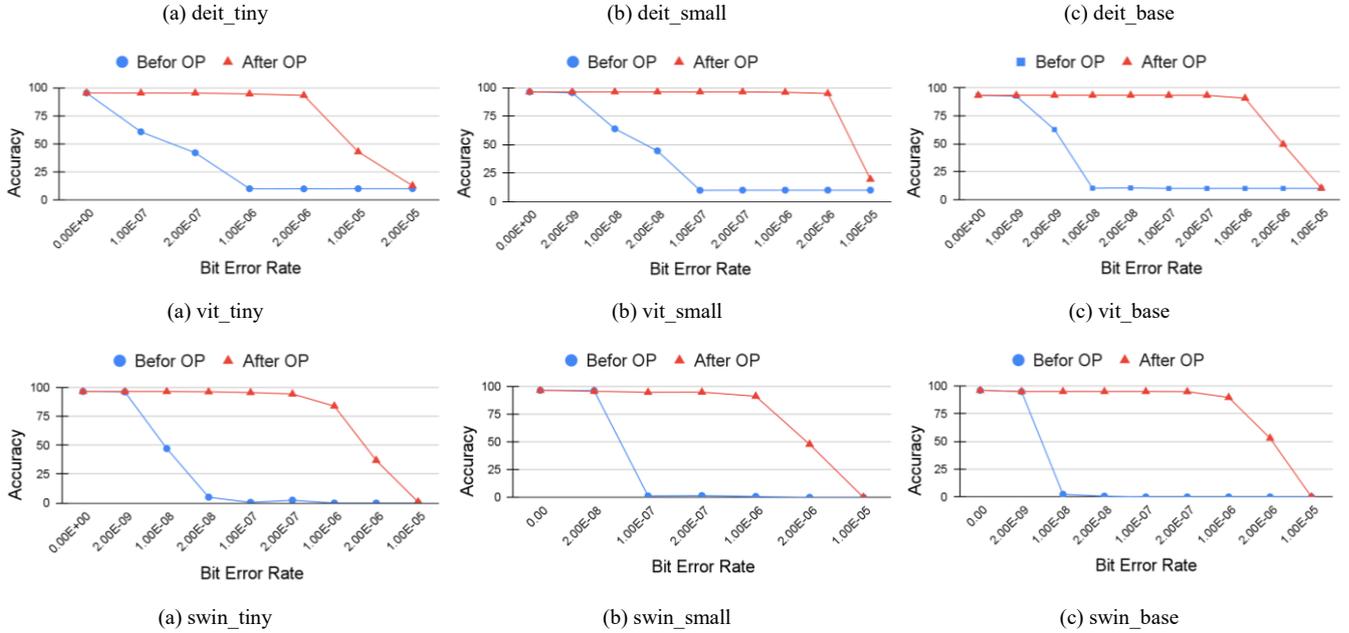

Fig. 5. Impact of OP on Accuracy in Different Models.

The evaluation also shows that vulnerability to faults increases with model size, likely due to error propagation throughout the network, impacting overall accuracy. These findings emphasize the need for robust fault tolerance, especially for larger models in safety-critical applications.

### D. Detail Evaluation of the Accuracy Distribution

Fig. 6 provides a more detailed view of the accuracy distribution for our experiments on DeiT Tiny. In this figure, blue bars represent accuracy without OP, and red bars represent accuracy with OP. Notably, a significant shift toward higher reliability is observed at each BER level. At BER 2e-7, the blue bars show considerable variability, similar to the red bars at BER 2e-5, a critical threshold beyond which accuracy significantly declines. As shown in the figure, OP not only improves the average accuracy, as seen in Fig. 5, but also enhances the entire distribution of accuracy results, indicating its efficacy in improving the reliability of ViT models.

### E. Comparison with the state-of-the-art

In order to show the superiority of the proposed approach over the state-of-the-art fault tolerance technique approaches for ViTs, we compare the proposed approach with ALBERTA which is a checksum-based approach for GEMM layers in ViTs presented in [11]. Since both technique provide complete fault tolerance in BERs at the ground level (i.e. 1e-7 tp 1e-8), we compare the memory and computation overhead of these techniques for two ViT models (i.e. Vit_base and Deit_base), as repoted in Table II. According to the obtained results, the proposed method introduces zero memory overhead, whereas ALBERTA increases memory usage by 25%. For computational overhead, the proposed method depends entirely on XOR operations, whereas ALBERTA relies on more resource-intensive multiply and add operations. Assuming one addition equals two XORs and one multiplication equals 10–50 XORs, ALBERTA's computational overhead exceeds that of the proposed method by a factor of over 500. Thus, the proposed fault tolerance approach incurs zero memory overhead and less computation overhead compared to the state-of-the-art fault tolerance approach for ViTs.

TABLE II - COMPARISON OF THE PROPOSED APPROACH WITH ALBERTA [11] FOR TWO VIT MODELS IN TERMS OF MEMORY AND COMPUTATION OVERHEAD

| Network | Memory Overhead | | Computation Overhead | |
|---|---|---|---|---|
| | ALBERTA [11] | Ours | ALBERTA [11] | Ours |
| Vit_base | 25% | 0% | Multiply: 124.85 b Add: 126.66 m | XOR: 2.666 b |
| Deit_base | 25% | 0% | Multiply: 125.48 b Add: 127.10 m | XOR: 2.6598 b |

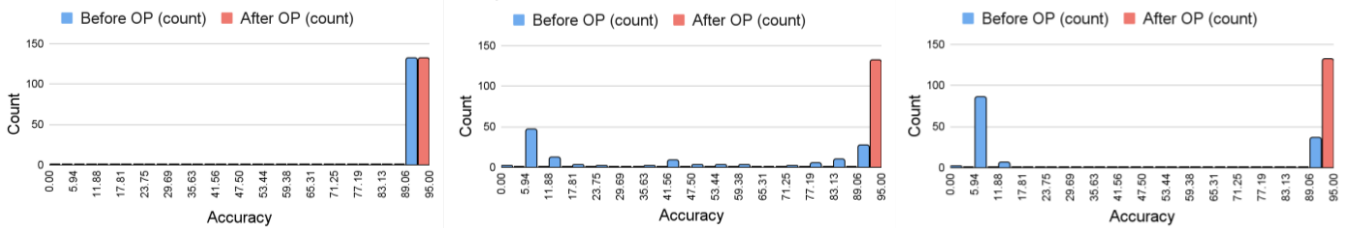

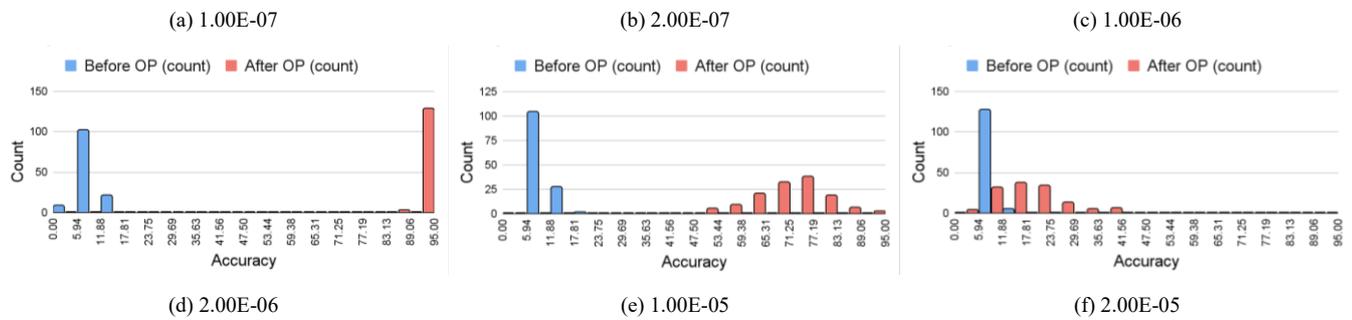

(a) 1.00E-07  (b) 2.00E-07  (c) 1.00E-06

(d) 2.00E-06  (e) 1.00E-05  (f) 2.00E-05

Fig. 6. Accuracy distribution for DeiT Tiny across BERs

## V. CONCLUSION

This paper introduces a zero-overhead fault tolerance technique to protect ViT parameters against bit-flip faults. The technique replaces each parameter's LSB with a parity bit, enabling error detection without additional memory overhead. Mitigation is achieved by zeroing out affected parameters, leveraging the fact that most ViT parameters are near zero. Experimental results indicate that this approach improves the robustness of parameters to bit-flips by up to three orders of magnitude. Furthermore, a detailed analysis reveals that not only does the average accuracy improve, but the entire accuracy distribution also shifts positively at each BER level. These findings highlight the proposed method's effectiveness as a zero-overhead fault tolerance solution, even at higher BER levels. Future work could explore adaptive error correction methods and assess the impact of this approach on quantized ViTs to further enhance reliability in safety-critical applications.

## REFERENCES

[1] Alexey Dosovitskiy, Lucas Beyer, Alexander Kolesnikov, Dirk Weissenborn, Xiaohua Zhai, Thomas Unterthiner, Mostafa Dehghani, Matthias Minderer, Georg Heigold, Sylvain Gelly, et al. An image is worth 16x16 words: Transformers for image recognition at scale. arXiv preprint arXiv:2010.11929, 2020.

[2] Ze Liu, Yutong Lin, Yue Cao, Han Hu, Yixuan Wei, Zheng Zhang, Stephen Lin, and Baining Guo. Swin transformer: Hierarchical vision transformer using shifted windows. CoRR, abs/2103.14030, 2021

[3] Salman Khan, Muzammal Naseer, Munawar Hayat, Syed Waqas Zamir, Fahad Shahbaz Khan, and Mubarak Shah. Transformers in vision: A survey. ACM computing surveys (CSUR), 54(10s):1–41, 2022.

[4] H. Touvron et al. Going deeper with image transformers. In IEEE/CVF International Conf. on Computer Vision, pages 32–42, 2021.

[5] Minhee Kang et al. "Vision Transformer for Detecting Critical Situations And Extracting Functional Scenario for Automated Vehicle Safety Assessment." SSRN Electronic Journal (2022).

[6] Zitao Chen, Guanpeng Li, and Karthik Pattabiraman. A low-cost fault corrector for deep neural networks through range restriction. In IEEE/IFIP International Conference on Dependable Systems and Networks (DSN), pages 1–13, 2021.

[7] Abdulrahman Mahmoud, Siva Kumar Sastry Hari, Christopher W Fletcher, Sarita V Adve, Charbel Sakr, Naresh R Shanbhag, Pavlo Molchanov, Michael B Sullivan, Timothy Tsai, and Stephen W Keckler. Optimizing selective protection for CNN resilience. In ISSRE, pages 127–138, 2021.

[8] Kuang-Hua Huang and Jacob A Abraham. Algorithm-based fault tolerance for matrix operations. IEEE transactions on computers, 100(6):518–528, 1984.

[9] Xue, Xinghua, et al. "ApproxABFT: Approximate algorithm-based fault tolerance for vision transformers." arXiv preprint arXiv:2302.10469 (2023).

[10] Ma, Kwondo, Chandramouli Amarnath, and Abhijit Chatterjee. "Error Resilient Transformers: A Novel Soft Error Vulnerability Guided Approach to Error Checking and Suppression." 2023 IEEE European Test Symposium (ETS). IEEE, 2023.

[11] Liu, Haoxuan, et al. "ALBERTA: ALgorithm-Based Error Resilience in Transformer Architectures." IEEE Open Journal of the Computer Society (2024).

[12] D . Zhou et al., "DeepViT: Towards deeper vision transformer," 2021, arXiv:2103.11886.

[13] S. Kim et al., "MATIC: Learning around errors for efficient low-voltage neural network accelerators." IEEE, Mar. 2018, pp. 1–6. [Online]. Available: http://ieeexplore.ieee.org/document/8341970/

[14] M. Sabbagh et al., "Evaluating fault resiliency of compressed deep neural networks," in 2019 IEEE International Conference on Embedded Software and Systems (ICESS), 2019, pp. 1–7.

[15] Burel, Stéphane, Adrian Evans, and Lorena Anghel. "Zero-overhead protection for cnn weights." 2021 IEEE International Symposium on Defect and Fault Tolerance in VLSI and Nanotechnology Systems (DFT). IEEE, 2021.

[16] Xue, Xinghua, et al. "Soft error reliability analysis of vision transformers." IEEE Transactions on Very Large Scale Integration (VLSI) Systems (2023).